\documentclass[11pt]{article}
\usepackage[margin=1in]{geometry}
\usepackage{graphicx}
\usepackage[normalem]{ulem}
\usepackage{tikz-cd}

\usepackage{adjustbox}

\usepackage{amsmath,amssymb}
\usepackage{hyperref}
\hypersetup{
colorlinks = true,
linkcolor = violet,
citecolor = violet,
urlcolor = violet
}

\title{Argumentative Topology: Finding Loop(holes) in Logic}

\title{Argumentative Topology: Finding Loop(holes) in Logic}

\author{
  Sarah Tymochko$^{1,2}$, Zachary New$^{2}$, Lucius Bynum$^{2}$, Emilie Purvine$^{2}$,\\ Timothy Doster$^{2}$, Julien Chaput$^{2,3}$, Tegan Emerson$^{2}$
}

\date{
$^1$Dept. of Computational Mathematics, Science and Engineering, Michigan State University \\
$^2$Pacific Northwest National Laboratory \\
$^3$Department of Geological Sciences, University of Texas at El Paso \\
}

\begin{document}

\maketitle

\begin{abstract}
  Advances in natural language processing have resulted in increased capabilities with respect to multiple tasks. One of the possible causes of the observed performance gains is the introduction of increasingly sophisticated text representations. While many of the new word embedding techniques can be shown to capture particular notions of sentiment or associative structures, we explore the ability of two different word embeddings to uncover or capture the notion of logical shape in text. To this end we present a novel framework that we call \textit{Topological Word Embeddings} which leverages mathematical techniques in dynamical system analysis and data driven shape extraction (i.e. topological data analysis). In this preliminary work we show that using a topological delay embedding we are able to capture and extract a different, shape-based notion of logic aimed at answering the question ``Can we find a circle in a circular argument?''
\end{abstract}

\section{Introduction}
\label{sec:intro}
Development of increasingly complex word embeddings have enabled algorithmic performance gains in tasks including, but not limited to, machine translation, author attribution, text summarization, argument mining, and question answering, e.g. \cite{cabrio2018argumentmining,devlin2018bert,stamatatos2016authorship,radford2019language}. Movement beyond the still widely used Bag-of-Words or Term Frequency-Inverse Document Frequency (TF-IDF) models of language was inspired by a desire to account for word-order dependencies and more aptly capture human notions of sentiment. For example, one of the many appeals of transformer network-based embeddings of language is that they capture human interpretable relationships like capital:country and gender:royal title \cite{mikolov2013efficient,mikolov2013distributed}.

What we present here is a novel framework for considering the capacity of different embeddings to model shape-based intuitions of logical argument structures. Others have also considered text data from a topological approach since the early 2010s, see \cite{zhu2013persistent}, \cite{michel2017does}, \cite{gholizadeh2018topological}, \cite{temcinas2018local}, and \cite{wagner2012computational}.

We were originally motivated by the question ``Why do we call a circular argument `circular'?'' A circular argument is one that logically loops back on itself. This intuitive definition of why the argument is circular actually has an analogous and mathematically precise definition from a topological perspective. Topology is the mathematical notion of shape and a topological circle can be abstractly defined as any shape which starts and loops back on itself (i.e. a circle, a square, and a triangle are all topologically a circle). What we present here is a novel combination of techniques from natural language processing, topological data analysis, and dynamical systems to understand the shape of argument structure.

First we will present a brief background in Section \ref{sec:back},  introducing the relevant computational tools (topological data analysis and time-delayed embeddings). In Section \ref{sec:WDE} we present topological word-delay embeddings for extraction of topology from text. Section \ref{sec:exp} contains our proof-of-concept examples and lays out our experimental design. Finally, we conclude with a discussion of our initial results and future directions in Sections~\ref{sec:results} and~\ref{sec:conc}.

\section{Materials and Methods}
\label{sec:back}
\subsection{Topological Data Analysis}
\label{subsec:TDA}
Topological data analysis (TDA) is built on the principle that data has shape, and that shape has meaning \cite{carlsson2009topology}. In order to understand this statement we must understand what ``shape'' means, how it is represented mathematically, and how that representation can be used to find meaning. For our purposes we will use the concepts of homology and persistent homology as mathematical representations of shape.
For a detailed description of these concepts see \cite{carlsson2009topology,ghrist2008barcodes}. In this paper we will provide the intuition behind the concepts and leave the details to the references.

The \textbf{homology} of a topological space (e.g., a torus, sphere, or arbitrary simplicial complex) identifies and quantifies the space's holes at each dimension. In dimension 0 holes are connected components of the space, in dimension 1 they are cycles, and in dimension 2 they are enclosed voids like the inside of a hollow sphere.
Homology is useful to quantify the holes present in a single space. But in our setting we will be dealing with point cloud data, a collection of points in high dimensional space. While this collection of points technically is itself a simplicial complex, its topology is uninteresting. Instead we study a \emph{filtration} of spaces derived from pairwise distances in the point cloud. This allows us to compute the homology at each scale and identify which features \textit{persist} over multiple scales. This framework is called \textbf{persistent homology}.

There are multiple ways for a user to perform a filtration, largely dependent on the type of data being processed. Here, given that our data is in the form of an embedded point cloud we use a distance-based filtration and extract the homology based on the construction of the \textbf{Vietoris-Rips} complex at each distance threshold \cite{carlsson2009topology}. In this way the persistent homology of a point cloud $X$ is taken as the collection of $(b, d)$ pairs ($b$ = birth, $d$ = death) over all holes present in the filtration. Each $(b,d)$ pair corresponds to the distance thresholds at which a single hole first appeared and then ceased in the corresponding complexes, respectively.  A common representation of the persistent homology is a \textbf{persistence diagram} or a plot of this multiset of $(b, d)$ points in the $\mathbb{R}^2$ plane.

\subsection{Time-Delayed Embeddings}
\label{subsec:TDE}
Time-Delay Embedding (TDE) is a technique for time-domain signal analysis based on the seminal work of Takens \cite{Takens:81} and Whitney \cite{Whitney:36}. The underlying assumption of TDE is that, given a dynamical system describing a physical process, some measured aspect of the system is a time-varying statistic from which certain uniquely identifying dynamic invariants may be extracted given a ``proper'' embedding of the observed statistic.

The output of a TDE is a cloud of $D$-dimensional points constructed from delayed copies of the observed $1$-dimensional time series. 
Such an embedding is constructed based on two parameters, an embedding dimension $D$, and a time-delay, $\tau$, and an observed time series, $\vec{z}=\{z_{n}\}_{n=1}^{N}$, resulting in a point cloud, $Z=\{(z_{n}, z_{n+\tau}, \cdots, z_{n+(D-1)\tau}) \}_{n=1}^{N-(D-1)\tau}$.
The parameters can be selected using algorithms such as false nearest neighbors \cite{Kennel:92} for the embedding dimension, and mutual information \cite{Fraser:86} or autocorrelation \cite{Kantz2004} for the time-delay. 
Though some dynamic information may be lost due to measuring only a single statistic, Takens' theorem guarantees that certain dynamic invariants describing the topology of the underlying manifold on which the dynamics evolve will still be preserved and observable via the delay embedding process.

\subsection{Topological Word-Delay Embeddings}
\label{sec:WDE}

\begin{figure}
    \centering
    \includegraphics[width=\textwidth]{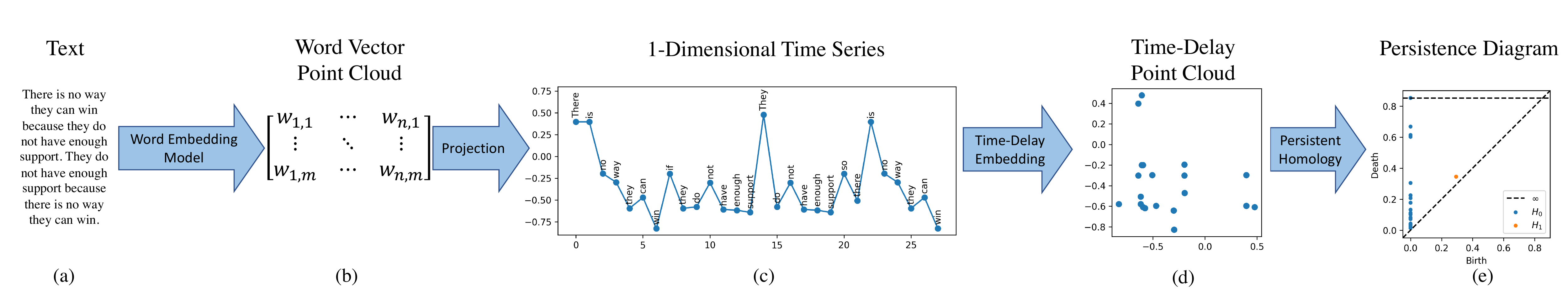}
    \caption{This figure illustrates the pipeline of our proposed Topological Word-Delay Embedding. }
    \label{fig:pipeline}
\end{figure}

We will now present our proposed approach for exploring the topology of logical structures that leverages techniques from dynamical systems analysis (namely time-delayed embeddings) and topological data analysis. Our approach is driven by the following thought experiment: consider an argument, presented in text, as a time evolving dynamical system that is a function of several inputs. 
Different types of argument structures could be described by the attractors of the corresponding dynamical systems. Here individual arguments would be analagous to individual trajectories of a dynamical system given an initial condition while a class of arguments share the same attractor structure and properties. From a dynamical systems perspective we know that under the necessary assumptions we can recover dynamic invariants of the attractor through TDEs. Included in the set of invariants which are preserved are those of a topological nature.

In order to leverage techniques in both TDEs and topology we implement the processing pipeline illustrated in Figure \ref{fig:pipeline}. We begin by embedding our text into a word-vector point cloud using a pretrained model, see (a)$\rightarrow$(b). Next we construct a $1-$dimensional time series by projecting each embedded word onto a fixed subspace (i.e. computing the dot product of each word vector in the text with a fixed random vector), see (b)$\rightarrow$(c). This one dimensional time series can then be embedded using TDEs and the resulting point cloud can be analyzed using persistent homology to yield a persistence diagram, see (c)$\rightarrow$(e). This process is shown in Figure \ref{fig:pipeline}. The word-delay point cloud shown in (d) was produced using embedding dimension $D=2$ (selected using a false nearest neighbors algorithm) and delay $\tau=2$ (determined using autocorrelation). These are the algorithms we will use in our experiments to select the parameters.

\subsection{Experimental Design} \label{sec:exp}
For our exploration we will be focusing on the following three text samples:
\begin{enumerate} \setlength{\itemsep}{0pt}
    \item A short valid argument: ``There is no way they can win if they do not have enough support. They do not have enough support, so there is no way they can win.''
    \item  A short invalid argument with circular logic: ``There is no way they can win because they do not have enough support. They do not have enough support because there is no way they can win.''
    \item Randomly generated text with a regular grammatical structure (using \url{randomwordgenerator.com}): ``The body may perhaps compensates for the loss of a true metaphysics. Yeah, I think it's a good environment for learning English. Wednesday is hump day, but has anyone asked the camel...
\end{enumerate}

We have chosen the above examples for a variety of reasons. The first two are very similar in terms of the words they are comprised of but the first is valid (``if premise then conclusion; premise is true thus conclusion is true'') while the second is invalid and circular (``conclusion is true because premise is true; premise is true because conclusion is true''). Finally, we compare to a random dialogue which is grammatically correct but doesn't inspire an obvious notion of shape.

Word embeddings are performed using two pretrained models: Word2Vec trained on the Google News dataset \cite{mikolov2013distributed} and GloVe trained on Common Crawl \cite{pennington2014glove}. 
We compute the persistence diagrams of the topological word-delay embeddings based on both of these models.
For comparison, we will also compute persistence diagrams of the word-vector point clouds formed by embedding all the words in the example texts using the indicated model. This allows for a comparison of whether the topological methods are able to extract more structure from the TDEs than just the word embeddings alone. In our preliminary work we also considered using FastText for embedding. Interestingly, FastText was not able to detect significant circular structure in either the word-vector clouds or the word-delay embedded clouds for any of the text examples.

\section{Results}
\label{sec:results}

\begin{figure}
    \centering
    \includegraphics[width=0.7\textwidth]{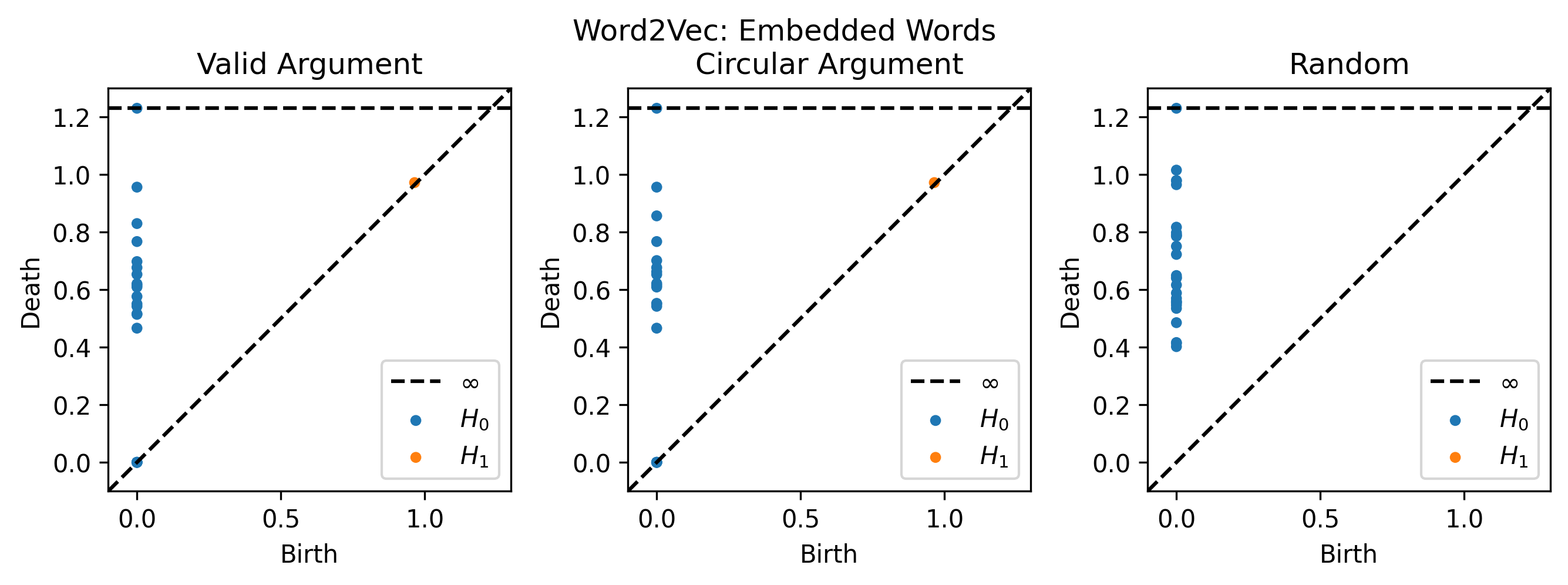} \\
    \includegraphics[width=0.7\textwidth]{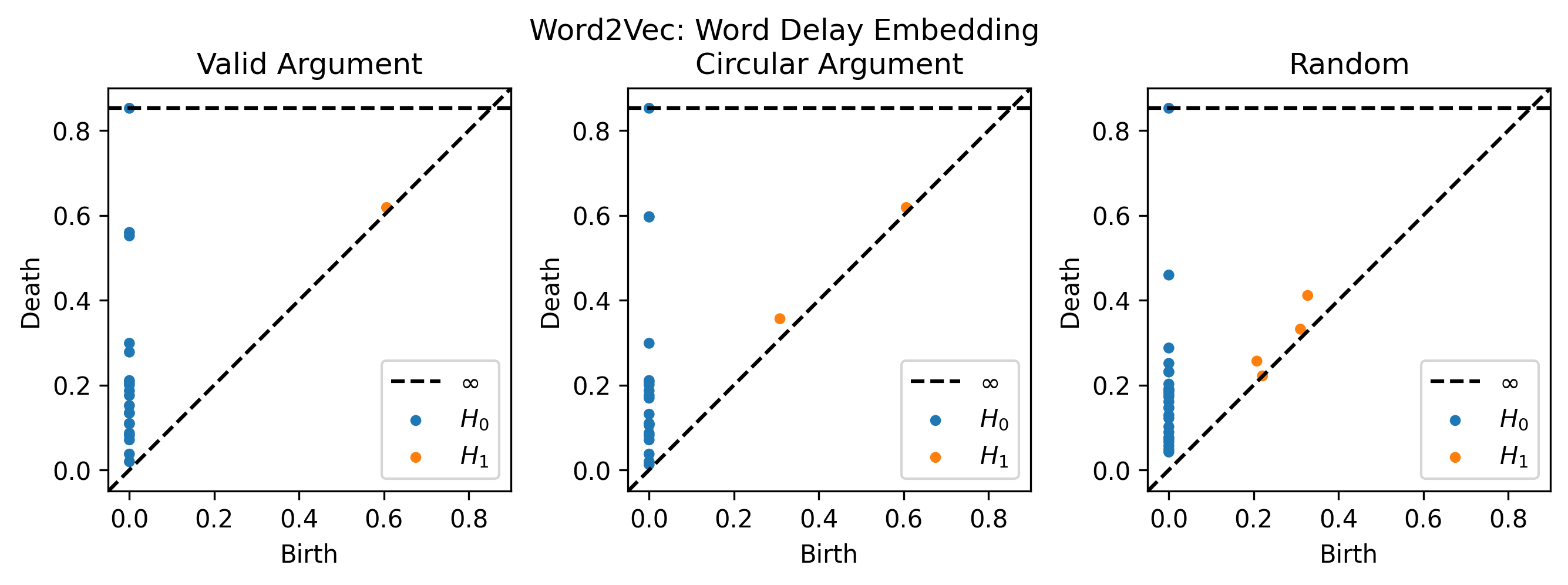}
    \caption{
    First row: Persistence diagrams computed based on vectors of embedded words using Word2Vec; second row: persistence diagrams based on WDE technique using Word2Vec.}
    \label{fig:PDs-word2vec}
\end{figure}

\begin{figure}
    \centering
    \includegraphics[width=0.7\textwidth]{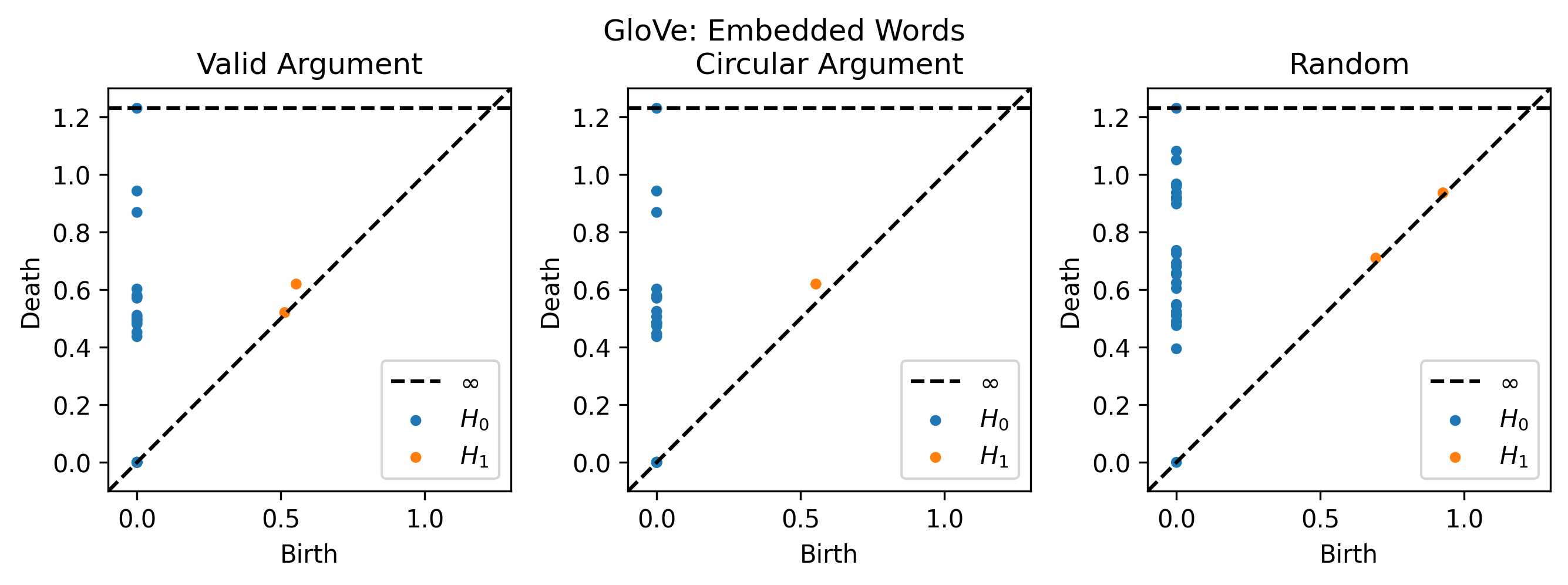} \\
    \includegraphics[width=0.7\textwidth]{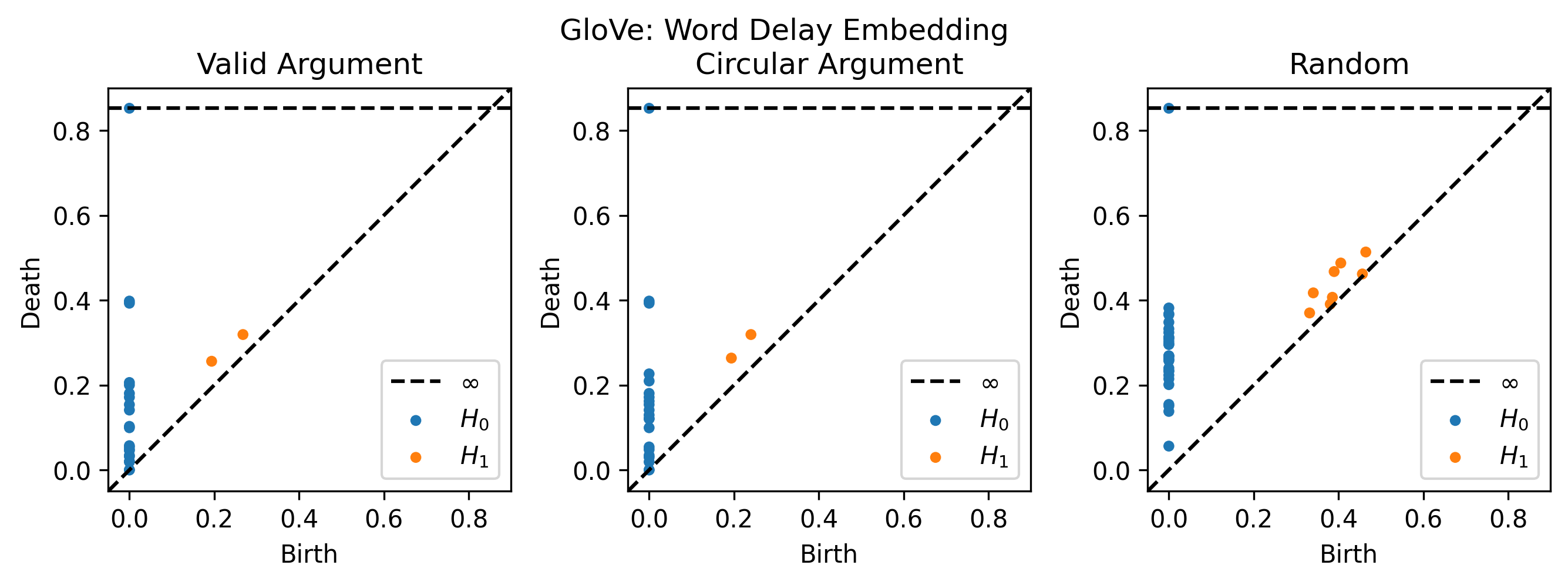}
    \caption{
   First row: Persistence diagrams computed based on vectors of embedded words using GloVe; second row: persistence diagrams based on WDE technique using GloVe.}
    \label{fig:PDs-glove}
\end{figure}

Our main results are shown in Figures~\ref{fig:PDs-word2vec} and \ref{fig:PDs-glove}.
In these figures we show the resulting persistence diagrams computed on the embedded words (first row in each figure) and the WDEs (second row in each figure) for both of the word embedding models.
We want to compare the persistence diagrams computed on the embedded words with those computed on the WDEs.
Specifically, we want to focus on 1-dimensional features, represented as orange points.

Looking at Fig.~\ref{fig:PDs-word2vec} using Word2Vec for embedding, simply computing persistence on the embedded word vectors alone reveals little topological information.
The valid argument and circular argument have only one 1-dimensional persistence point very close to the diagonal. Typically points near the diagonal are considered to be caused by noise while those farther from the diagonal represent significant features.
However, looking at the persistence diagrams computed on the WDEs, the persistence diagram for the circular argument has one additional, more substantial 1-dimensional point, while the random text has a few circular features.

Comparing the results in Fig. ~\ref{fig:PDs-glove} using GloVe for embedding, we see the embedded word vectors produce a 1-dimensional feature for both the valid and circular argument, while any 1-dimensional structure in the random text is noise.
The persistence diagrams for the WDEs pick up two circular features for each of the valid and circular arguments, however the circular features in the circular argument are slightly further from the diagonal.
The WDEs also produce numerous 1-dimensional features for the random text.



\section{Discussion}
\label{sec:conc}
What we have presented here is a proof-of-concept work aimed at beginning to answer the questions ``Can we find a circle in a circular argument?'' and ``What can we learn from the shape of logical constructs?"
We have shown that one can in fact detect the presence of topological circles in circular logic. Additionally, we have observed that our topological word embedding approach captures features of the data not detected studying word embeddings alone.

In ongoing and future work we will focus on quantitatively answering our second motivational question. To this end we are developing theory to provide mathematically principled support for the generation of topological WDEs.
Moreover, as we seek to leverage these topological insights for the identification of logical fallacies, we will be applying our approach to corpuses of text.
As we extend this method beyond the simple examples here and into larger processing pipelines we will need to be able to make quantitative comparisons between the topological features.
Towards this end we will be employing persistence images \cite{adams2017persistence}, a popular persistence diagram vectorization technique, and exploring novel techniques for the fusion of information from multiple homological dimensions.

\bibliographystyle{plain}
\bibliography{ArgumentativeTopology.bib}

\begin{thebibliography}{10}

\bibitem{adams2017persistence}
Henry Adams, Tegan Emerson, Michael Kirby, Rachel Neville, Chris Peterson,
  Patrick Shipman, Sofya Chepushtanova, Eric Hanson, Francis Motta, and Lori
  Ziegelmeier.
\newblock Persistence images: A stable vector representation of persistent
  homology.
\newblock {\em The Journal of Machine Learning Research}, 18(1):218--252, 2017.

\bibitem{cabrio2018argumentmining}
Elena Cabrio and Serena Villata.
\newblock Five years of argument mining: a data-driven analysis.
\newblock In {\em IJCAI}, pages 5427--5433, 2018.

\bibitem{carlsson2009topology}
Gunnar Carlsson.
\newblock Topology and data.
\newblock {\em Bulletin of the American Mathematical Society}, 46(2):255--308,
  2009.

\bibitem{devlin2018bert}
Jacob Devlin, Ming-Wei Chang, Kenton Lee, and Kristina Toutanova.
\newblock {BERT}: {P}re-training of deep bidirectional transformers for
  language understanding.
\newblock {\em arXiv preprint arXiv:1810.04805}, 2018.

\bibitem{Fraser:86}
Andrew~M. Fraser and Harry .~L. Swinney.
\newblock Independent coordinates for strange attractors from mutual
  information.
\newblock {\em Physical Review}, 33:1134--1140, 1986.

\bibitem{gholizadeh2018topological}
Shafie Gholizadeh, Armin Seyeditabari, and Wlodek Zadrozny.
\newblock Topological signature of 19th century novelists: Persistent homology
  in text mining.
\newblock {\em Big Data and Cognitive Computing}, 2(4):33, 2018.

\bibitem{ghrist2008barcodes}
Robert Ghrist.
\newblock Barcodes: {T}he persistent topology of data.
\newblock {\em Bulletin of the American Mathematical Society}, 45(1):61--75,
  2008.

\bibitem{Kantz2004}
Holger Kantz and Thomas Schreiber.
\newblock {\em Nonlinear time series analysis}, volume~7.
\newblock Cambridge university press, 2004.

\bibitem{Kennel:92}
Matthew~B. Kennel, Reggie Brown, and Henry D.~I. Abarbanel.
\newblock Determining embedding dimension for phase-space reconstruction using
  a geometrical construction.
\newblock {\em Physical Review A}, 45:3403--3411, 1992.

\bibitem{michel2017does}
Paul Michel, Abhilasha Ravichander, and Shruti Rijhwani.
\newblock Does the geometry of word embeddings help document classification?
  {A} case study on persistent homology based representations.
\newblock {\em arXiv preprint arXiv:1705.10900}, 2017.

\bibitem{mikolov2013efficient}
Tomas Mikolov, Kai Chen, Greg Corrado, and Jeffrey Dean.
\newblock Efficient estimation of word representations in vector space.
\newblock {\em arXiv preprint arXiv:1301.3781}, 2013.

\bibitem{mikolov2013distributed}
Tomas Mikolov, Ilya Sutskever, Kai Chen, Greg~S Corrado, and Jeff Dean.
\newblock Distributed representations of words and phrases and their
  compositionality.
\newblock In {\em Advances in neural information processing systems}, pages
  3111--3119, 2013.

\bibitem{pennington2014glove}
Jeffrey Pennington, Richard Socher, and Christopher~D Manning.
\newblock Glove: Global vectors for word representation.
\newblock In {\em Proceedings of the 2014 conference on empirical methods in
  natural language processing (EMNLP)}, pages 1532--1543, 2014.

\bibitem{radford2019language}
Alec Radford, Jeff Wu, Rewon Child, David Luan, Dario Amodei, and Ilya
  Sutskever.
\newblock Language models are unsupervised multitask learners.
\newblock 2019.

\bibitem{stamatatos2016authorship}
Efstathios Stamatatos.
\newblock Authorship verification: {A} review of recent advances.
\newblock {\em Research in Computing Science}, 123:9--25, 2016.

\bibitem{Takens:81}
Florius Takens.
\newblock Detecting strange attractors in turbulence.
\newblock In D.A. Rand and L.-S Young, editors, {\em Dynamical Systems and
  Turbulence}, volume 898 of {\em Lecture Notes in Mathematics}, pages 366--81.
  Springer-Verlag, New York, 1981.

\bibitem{temcinas2018local}
Tadas Temcinas.
\newblock Local homology of word embeddings.
\newblock {\em arXiv preprint arXiv:1810.10136}, 2018.

\bibitem{wagner2012computational}
Hubert Wagner, Pawe{\l} D{\l}otko, and Marian Mrozek.
\newblock Computational topology in text mining.
\newblock In {\em Computational Topology in Image Context}, pages 68--78.
  Springer, 2012.

\bibitem{Whitney:36}
Hassler Whitney.
\newblock Differentiable manifolds.
\newblock {\em Annals of Mathematics}, pages 645--680, 1936.

\bibitem{zhu2013persistent}
Xiaojin Zhu.
\newblock Persistent homology: An introduction and a new text representation
  for natural language processing.
\newblock In {\em Twenty-Third International Joint Conference on Artificial
  Intelligence}, 2013.

\end{thebibliography}

\end{document}